\documentclass[runningheads]{llncs}
\usepackage{graphicx}
\usepackage{comment}
\usepackage{amsmath,amssymb} 
\usepackage{color}

\usepackage{my_pkg}

\begin{document}
\pagestyle{headings}
\mainmatter
\def\ECCVSubNumber{2595}  

\title{Learning to Generate Novel Domains for Domain Generalization} 

\titlerunning{Learning to Generate Novel Domains for Domain Generalization}
%
\author{
Kaiyang Zhou\inst{1} \and
Yongxin Yang\inst{1} \and
Timothy Hospedales\inst{2,3} \and
Tao Xiang\inst{1,3}
}
\authorrunning{Zhou et al.}
%
\institute{University of Surrey \\
\email{\{k.zhou, yongxin.yang, t.xiang\}@surrey.ac.uk}
\and
University of Edinburgh \\
\email{t.hospedales@ed.ac.uk} \and
Samsung AI Center, Cambridge
}
\maketitle

\begin{abstract}
This paper focuses on domain generalization (DG), the task of learning from multiple source domains a model that generalizes well to unseen domains. A main challenge for DG is that the available source domains often exhibit limited diversity, hampering the model's ability to learn to generalize. We therefore employ a data generator to synthesize data from pseudo-novel domains to augment the source domains. This explicitly increases the diversity of available training domains and leads to a more generalizable model. To train the generator, we model the distribution divergence between source and synthesized pseudo-novel domains using optimal transport, and maximize the divergence. To ensure that semantics are preserved in the synthesized data, we further impose cycle-consistency and classification losses on the generator. Our method, L2A-OT (Learning to Augment by Optimal Transport) outperforms current state-of-the-art DG methods on four benchmark datasets.
\end{abstract}

\section{Introduction} \label{sec:intro}
Humans effortlessly generalize prior knowledge to novel scenarios, a capability that machines still struggle to reproduce. Typically, machine-learning models perform poorly when deployed on test data with a different data distribution than the training data, which is known as the domain shift problem~\cite{zhou2021domain,moreno2012unifying}. One line of research towards alleviating the domain shift problem is unsupervised domain adaptation (UDA), which exploits unlabeled target domain data for model adaptation~\cite{ganin2015unsupervised,long2015learning,hoffman2018cycada,Peng_2019_ICCV,Xu_2019_ICCV,Saito_2019_ICCV}. Although UDA methods avoid costly data annotation processes from target domains, data collection and per-domain model updates are still required. Meanwhile, UDA's assumption that target data can be collected in advance is not always met in practice~\cite{muandet2013domain,dou2019domain}. This motivates another line of research, namely domain generalization (DG)~\cite{zhou2021domain,muandet2013domain,ghifary2015domain,ghifary2017scatter,balaji2018metareg,cvpr19JiGen,dou2019domain}, which is the main focus in this paper.

\begin{figure}[t]
    \centering
    \includegraphics[width=.8\textwidth]{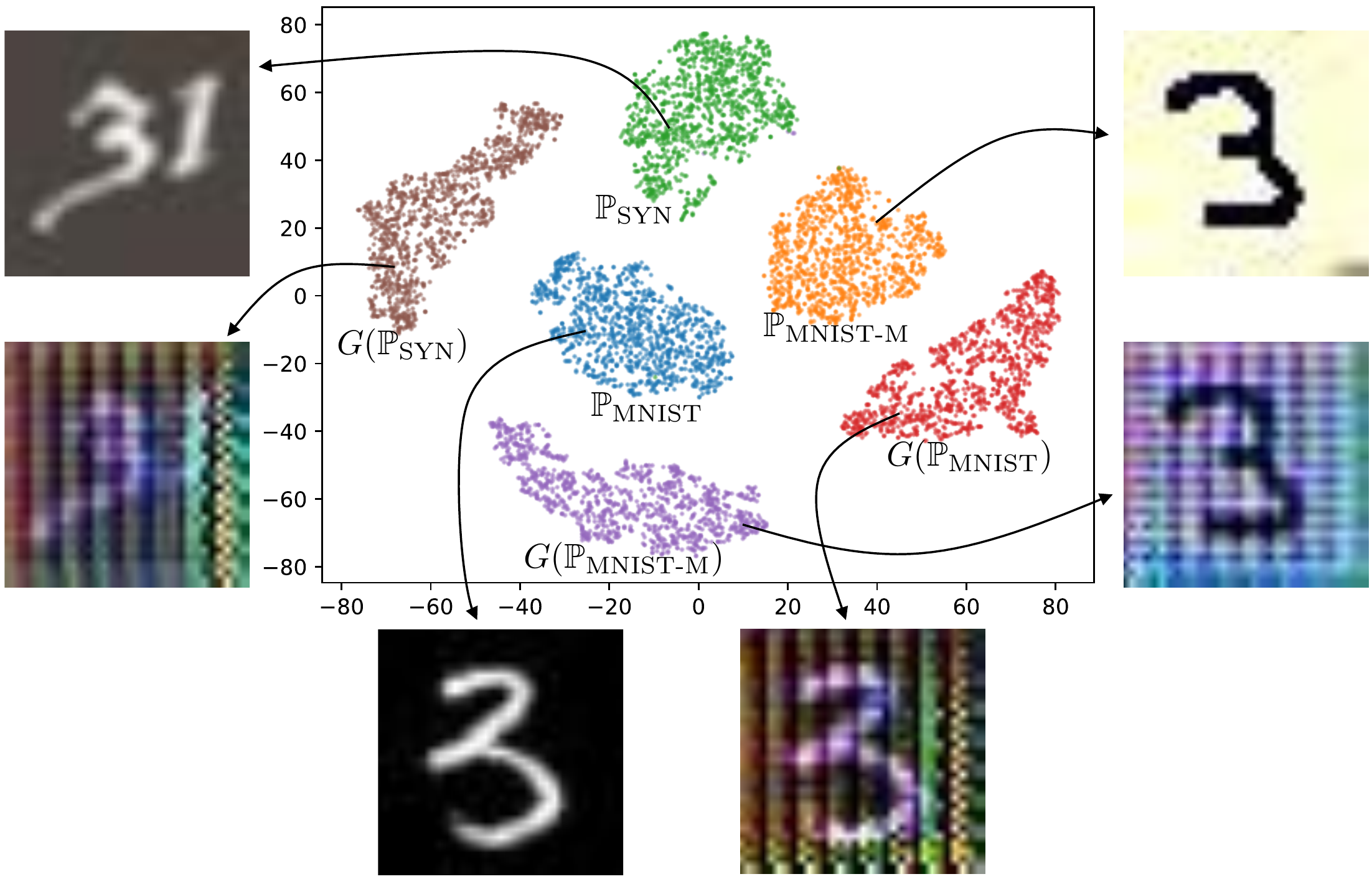}
    \caption{Motivation of our approach. We improve generalization by increasing the diversity of training domains by learning a generator network $G$ to map images of a source distribution, e.g., $\mathbb{P}_{\text{MNIST}}$, to a novel distribution, i.e. $G(\mathbb{P}_{\text{MNIST}})$. We then combine both source and novel domains for model learning.}
    \label{fig:main_idea}
\end{figure}

DG methods aim to learn models capable of good direct generalization to unseen target domains without data collection or model updating~\cite{muandet2013domain}. They usually, but not always~\cite{volpi2018generalizing}, leverage multiple source domains to train a generalizable model. Most existing DG methods focus on aligning available source domains~\cite{motiian2017unified,ghifary2017scatter,gan2016learning,ghifary2015domain,li2018ciddg,li2018mmdaae}, which is mainly inspired by UDA methods that seek to minimize the divergence between source data and unlabeled target data~\cite{ganin2016domain,tzeng2017adversarial}. As proved in~\cite{ben2010theory}, minimizing the domain divergence can lead to a smaller target error in the UDA setting. However, since DG methods focus on aligning source domains and do not have access to the target data, this theoretical proof does not apply to the DG setting. Recently, meta-learning has been exploited for DG where the key idea is to simulate domain shift by splitting the training data into meta-train and meta-test sets with non-overlapping domains~\cite{li2018learning,balaji2018metareg,feature_critic,li2019episodic,dou2019domain}. During learning, models are optimized on the meta-train domains in a way that the error is reduced on the meta-test domains. Nevertheless, similar to the alignment-based methods, meta-learning optimizes for reducing the domain gap among source domains, and thus still has the risk of overfitting to seen domains.

In this paper, we address DG from a different perspective, i.e., the most straightforward way to improve model generalization is increasing the diversity of available source domains~\cite{DomainRandomization} (see Fig.~\ref{fig:main_idea}). To this end, we propose \emph{L2A-OT} (\emph{Learning to Augment by Optimal Transport}). The core idea is to learn a conditional generator network that maps source domain images to pseudo-novel domains, and then combine both source and pseudo-novel domain images for training the actual task model. To train the generator, we \emph{maximize} the distance between source domains and the generated pseudo-novel domains, as measured by optimal transport (OT)~\cite{peyre2019computational}. This leads to the generated images having a very different distribution from the source domains (Fig.~\ref{fig:main_idea}). However, this objective alone does not guarantee that the semantic content of the generated images is preserved. Therefore, we further impose two losses on the generator, namely a cycle-consistency loss~\cite{CycleGAN} and a classification loss, for maintaining the structural and semantic consistency respectively.

Our contributions are as follows. \textbf{(1)} For the first time, DG is tackled from a perspective of pseudo-novel domain synthesis. \textbf{(2)} A novel image generator is formulated which differs from existing generators in the objective (synthesizing pseudo-novel domain images vs.~natural photo images). More importantly it has  a unique OT-based formulation of objective functions that allow the generator to explore novel domain space and generate diverse data with distributions different from any of the original source domains. We evaluate L2A-OT on three homogeneous DG benchmark datasets\footnote{Following~\cite{feature_critic}, homogeneous DG shares the same label space between training and test data while heterogeneous DG has disjoint label space.} including digit recognition~\cite{lecun1998mnist,ganin2015unsupervised,netzer2011svhn}, PACS~\cite{li2017deeper} and Office-Home~\cite{office_home} and a heterogeneous DG task in the form of cross-domain person re-identification (re-ID)~\cite{zhong2018gen,zhong2019camstyle,liu2019adaptive,zhou2019learning,jin2020style}. The results show that L2A-OT surpasses the current state-of-the-art on all datasets.

\section{Related Work} \label{sec:related_work}
\paragraph{Domain generalization.}
Many DG methods are based on the idea of domain alignment popularized from the UDA literature~\cite{ganin2015unsupervised}, with a goal to learn a domain-invariant representation by minimizing the domain discrepancy between sources~\cite{motiian2017unified,ghifary2017scatter,gan2016learning,ghifary2015domain,li2018ciddg,li2018mmdaae}. As mentioned earlier, aligning domain distributions is mainly motivated by the theory~\cite{ben2010theory} developed for UDA, which does not apply to DG due to the absence of target data. Therefore, the models learned with domain alignment risk overfitting to source domains and as a result generalize poorly to unseen domains. In recent years, meta-learning~\cite{hospedales2020meta} has seen increasing interest for DG where the objective is to expose a model to domain shift during training. This can be achieved by dividing source domains into meta-train and meta-test sets without overlapping, and training a model on the meta-train set such that the error on the meta-test set is reduced~\cite{li2018learning,balaji2018metareg,dou2019domain}. Similar to domain alignment methods, meta-learning methods still risk overfitting since the training data remains unchanged. Moreover, these methods work at feature-level, which is difficult for diagnosis and lacks visual interpretation. We refer readers to~\cite{zhou2021domain} for a comprehensive DG survey.

Most related to our work are data augmentation methods, especially those based on adversarial gradients~\cite{shankar2018generalizing,volpi2018generalizing}. For instance, \cite{shankar2018generalizing} proposed CrossGrad to perturb input images with adversarial gradients generated by a domain classifier. Different from adversarial gradient-based methods which only produce imperceptible and simple pixel-wise effects (due to the nature of adversarial attack~\cite{szegedy2014intriguing}), our approach \emph{learns} a full CNN generator to map source images to unseen domains and optimizes it via \emph{OT}-based distribution divergence to make the new domains as dissimilar as possible to source distributions.

\paragraph{Domain randomization.}
Our approach shares a similar high-level intuition with domain randomization (DR)~\cite{DomainRandomization}, which was originally introduced in the context of robotic learning to improve generalization from simulation to real world. DR aims to diversify the training domains by changing the color and texture of objects, background scenes, lighting conditions, etc. via a computer simulator~\cite{DomainRandomization}. Recently, DR has been successfully used in some computer vision applications, such as semantic segmentation~\cite{yue2019domain,zakharov2019deceptionnet} and vehicle detection for autonomous driving~\cite{prakash2019structured}. However, our approach is significantly different from the DR-based methods because we \emph{learn} a CNN generator network from real images rather than using programmatic simulators. Thus our method is more scalable to a wider range of image recognition tasks.

\paragraph{Image-to-image translation.}
Our work is also related to multi-domain image-to-image translation methods such as CycleGAN~\cite{CycleGAN} and StarGAN~\cite{StarGAN}, which use GAN losses~\cite{goodfellow2014generative} to generate realistic images and cycle-consistency losses~\cite{CycleGAN} to achieve translation without using paired training images. Our method is fundamentally different from CycleGAN/StarGAN in that our generator model is learned to map source images to \emph{unseen} domains rather than performing mapping between source domains as did in CycleGAN/StarGAN. We show by experiments that simply doing source-to-source mapping for data augmentation offers little help to DG (see Fig.~\ref{fig:ablation}a).

\section{Methodology} \label{sec:method}

\begin{figure}[t]
    \centering
    \includegraphics[width=\textwidth]{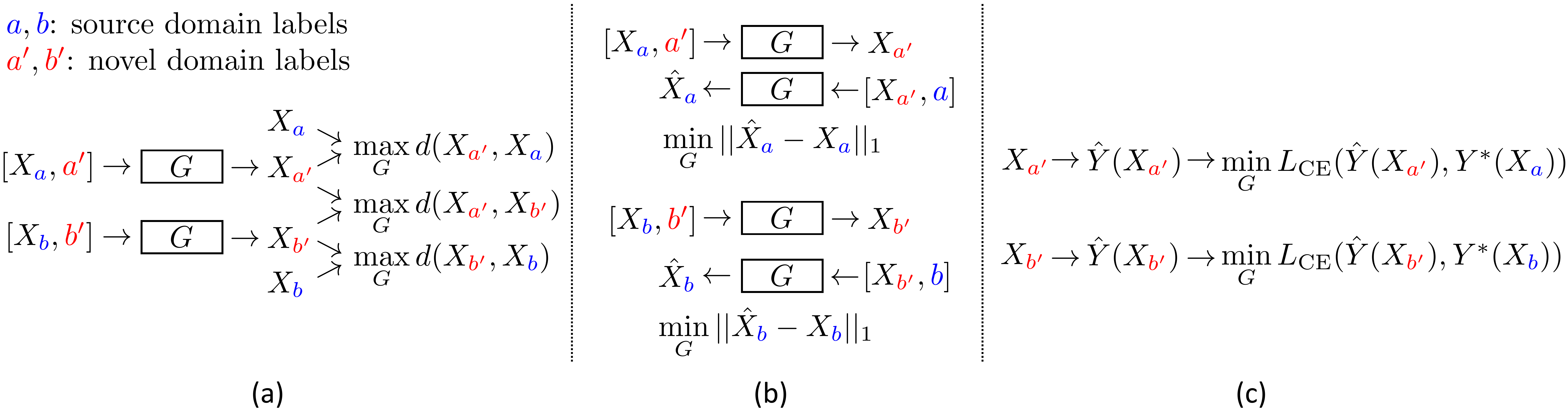}
    \caption{Overview of our approach. (a) The conditional generator network $G$ is learned to map input $X$ to novel domains whose distributions are drastically different from the source domains, while keeping the distance between the novel domains as far as possible. (b) A cycle-consistency loss is imposed on $G$ to maintain the structural consistency. (c) The cross-entropy loss is minimized with respect to $G$, using a pre-trained classifier $\hat{Y}$, for maintaining the semantic consistency.}
    \label{fig:train_g}
\end{figure}

\subsection{Generating Novel-Domain Data} \label{subsec:generate_novel_domain} 

\paragraph{Setup.}
We are provided with $K_s$ source domains with indices $D_s = \{ 1,2,...,K_s \}$. The goal is to learn a model which can generalize well on an unseen target domain. Without having access to the target data, we propose to improve the model's generalization by synthesizing novel data domains $D_n = \{ 1,2,...,K_n \}$ to augment the original source domains.

\paragraph{Conditional generator.}
We learn a conditional generator $G$ (see Sec.~\ref{subsec:design_g} for detailed architecture design), that maps a source distribution $\mathbb{P}_k$ with $k \in D_s$ to a novel distribution $\mathbb{P}_{\tilde{k}}$ with $\tilde{k} \in D_n$ by conditioning on the novel domain label $\tilde{k}$, i.e. $\mathbb{P}_{\tilde{k}}=G(\mathbb{P}_{k},\tilde{k})$. Here $\mathbb{P}$ denotes an empirical distribution rather than the real distribution, which is inaccessible. In practice, we use sampled mini-batches $X_k$ instead of the full empirical distribution $\mathbb{P}_k$. Therefore, the domain translation function is defined as:
\begin{equation} \label{eq:conditional_g}
X_{\tilde{k}} = G(X_k, \tilde{k}).
\end{equation}

\paragraph{Objective functions.}
For each training iteration, we randomly sample for each source domain $k$ a mini-batch $X_k$, which is transformed to a randomly selected novel domain $\tilde{k} \sim D_n$. The objective is to force the novel distribution to be as dissimilar as possible to any source distribution, thus creating new domains to augment the existing source domains. We have
\begin{equation} \label{eq:max_dom_diff}
\max_G L_{\mathrm{Novel}} = d(G(X_k, \tilde{k}), X_k),
\end{equation}
where $d(\cdot, \cdot)$ is a distribution divergence measure (its design will be detailed in Sec.~\ref{subsec:design_d}). Note that Eq.~\eqref{eq:max_dom_diff} will be summed over all source domains $k$, and each independently draws a novel domain label $\tilde{k}$.

In addition to maximizing the difference between source and novel distributions, we also maximize the difference between the generated novel distributions, i.e.
\begin{equation} \label{eq:diversity_term}
\max_G L_{\mathrm{Diversity}} = d(X_{\tilde{k}_1}, X_{\tilde{k}_2}),
\end{equation}
where $\tilde{k}_1, \tilde{k}_2 \in D_n$ and $\tilde{k}_1 \neq \tilde{k}_2$. Eq.~\eqref{eq:diversity_term} is summed over all possible pairs of novel distributions generated in one iteration. This diversity constraint diversifies the generated distributions, ensuring that the model benefits from generating $K_n>1$ novel distributions. It is analogous to the diversity term in some image generation tasks, such as style transfer~\cite{li2017diversified} where the pixel/feature difference between style-transferred instances is maximized. Differently, our formulation focuses on the divergence between data distributions. See Fig.~\ref{fig:train_g}a for a graphical illustration.

\subsection{Maintaining Semantic Consistency} \label{subsec:maintain_semantic}
The model so far is optimizing a powerful CNN generator $G$ for the novelty of the generated distribution (Eq.~\eqref{eq:max_dom_diff}~\&~\eqref{eq:diversity_term}). This produces diverse images, but may not preserve their semantic content.

\paragraph{Cycle-consistency loss.}
First, to guarantee structural consistency, we apply a cycle-consistency constraint~\cite{CycleGAN} to the generator,
\begin{equation} \label{eq:cycle}
\min_G L_{\mathrm{Cycle}} = || G(G(X_k, \tilde{k}), k) - X_k ||_1,
\end{equation}
where the outer $G$ aims to reconstruct the original $X_k$ given as input the domain-translated $G(X_k, \tilde{k})$ and the original domain label $k$. Both $G$'s in the cycle share the same parameters~\cite{StarGAN}. This is illustrated in Fig.~\ref{fig:train_g}b.

\paragraph{Cross-entropy loss.}
Second, to maintain the category label and thus enforce semantic consistency, we further require that the generated data $X_{\tilde{k}}$ is classified into the same category as the original data $X_k$, i.e.
\begin{equation} \label{eq:cross_entropy}
\min_G L_{\mathrm{CE}} ( \hat{Y}(X_{\tilde{k}}), Y^*(X_k) ),
\end{equation}
where $L_{\mathrm{CE}}$ denotes cross-entropy loss, $\hat{Y}(X_{\tilde{k}})$ the labels of $X_{\tilde{k}}$ predicted by a pretrained classifier and $Y^*(X_k)$ the ground-truth labels of $X_k$. This is illustrated in Fig.~\ref{fig:train_g}c.

\subsection{Training}
\paragraph{Generator training.}
The full objective for $G$ is the weighted combination of Eq.~\eqref{eq:max_dom_diff},~\eqref{eq:diversity_term},~\eqref{eq:cycle}, \&~\eqref{eq:cross_entropy},
\begin{equation} \label{eq:full_obj_g}
\begin{split}
\min_G L_G =
 & - \lambda_{\mathrm{Domain}} (L_{\mathrm{Novel}} + L_{\mathrm{Diversity}}) \\
 & + \lambda_{\mathrm{Cycle}} L_{\mathrm{Cycle}} + \lambda_{\mathrm{CE}} L_{\mathrm{CE}},
\end{split}
\end{equation}
where $\lambda_{\mathrm{Domain}}$, $\lambda_{\mathrm{Cycle}}$ and $\lambda_{\mathrm{CE}}$ are weighting hyper-parameters.

\paragraph{Task model training.}
The task model $F$ is trained from scratch using both the original data $X_k$ and the synthetic data $X_{\tilde{k}}$ generated as described above. The objective for $F$ is
\begin{equation} \label{eq:obj_f}
\min_F L_F = (1 - \alpha) L_{\mathrm{CE}} + \alpha \tilde{L}_{\mathrm{CE}},
\end{equation}
where $\alpha$ is a balancing weight, which is fixed to 0.5 throughout this paper; $L_{\mathrm{CE}}$ and $\tilde{L}_{\mathrm{CE}}$ are the cross-entropy losses computed using $X_{k}$ and $X_{\tilde{k}}$ respectively. The full training algorithm is shown in Alg.~\ref{alg:train_g_f}. Note that each source domain $k \in D_s$ will be assigned a unique novel domain $\tilde{k} \in D_n$ as target in each iteration. We set $K_n = K_s$ as default.

\begin{algorithm}[t]
\caption{Full training algorithm.}
\label{alg:train_g_f}
\footnotesize
\begin{algorithmic}[1] 
  \STATE \textbf{Require}: Source domain $D_s = \{ 1,2,...,K_s \}$, novel domain $D_n=\{ 1,2,...,K_n \}$, task model $F$, generator $G$, max iteration $T$, pre-trained classifier $\hat{Y}$.
  \FOR{$t = 1$ {\bf to} $T$}
    \STATE Sample a mini-batch $X_k$ from \emph{each} source domain $k \in D_s$.
    \STATE Generate $X_{\tilde{k}} = G(X_k, \tilde{k})$ with $\tilde{k} \sim D_n$ for \emph{each} source $k$.
    \STATE Compute $L_G$.
    \STATE Perform one step gradient update for $G$ using $\nabla_G L_G$.
    \STATE Compute $L_F$.
    \STATE Perform one step gradient update for $F$ using $\nabla_F L_F$.
    \ENDFOR
\end{algorithmic}
\end{algorithm}

\subsection{Design of Conditional Generator Network} \label{subsec:design_g}
\begin{figure}[t]
    \centering
    \includegraphics[width=\textwidth]{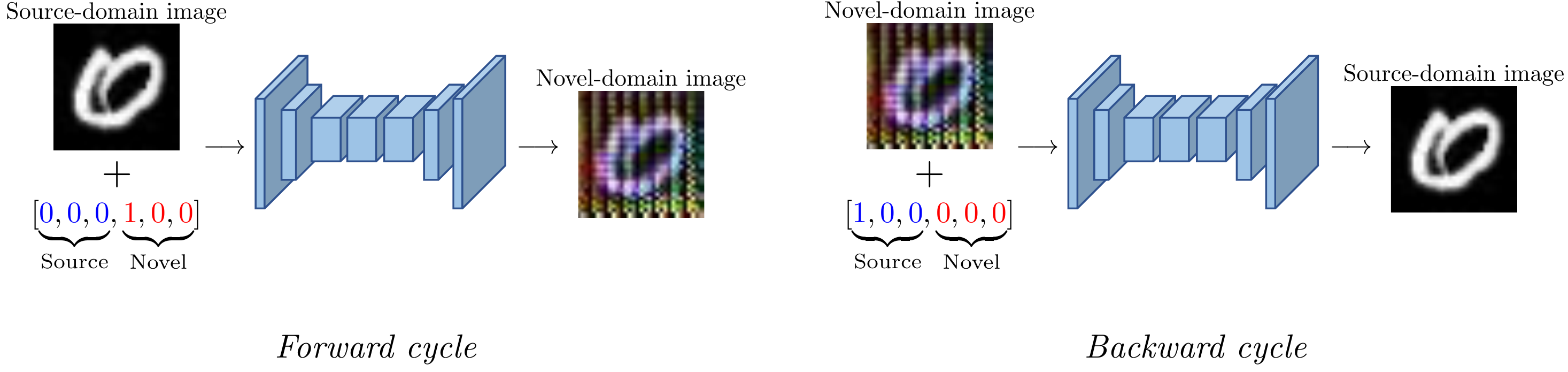}
    \caption{Architecture of the conditional generator network. Left and right images exemplify the forward cycle and backward cycle respectively in cycle-consistency.}
    \label{fig:g_architecture}
\end{figure}

Our generator model has a conv-deconv structure~\cite{CycleGAN,StarGAN} which is shown in Fig.~\ref{fig:g_architecture}. Specifically, the generator model consists of two down-sampling convolution layers with stride 2, two residual blocks~\cite{he2016deep} and two transposed convolution layers with stride 2 for up-sampling. Following StarGAN~\cite{StarGAN}, the domain indicator is encoded as a one-hot vector with length $K_s+K_n$ (see Fig.~\ref{fig:g_architecture}). During the forward pass, the one-hot vector is first spatially expanded and then concatenated with the image to form the input to $G$.

\paragraph{Discussion.}
Though the design of $G$ is similar to the StarGAN model, their learning objectives are totally different: We aim to generate images that are different from the existing source domain distributions while the StarGAN model is trained to generate images from the existing source domains. In the experiment part we justify that adding novel-domain data is much more effective than adding seen-domain data for DG (see Fig.~\ref{fig:ablation}a). Compared with the gradient-based perturbation method in~\cite{shankar2018generalizing}, our generator is allowed to model more sophisticated domain shift such as image style changes due to its learnable nature.

\subsection{Design of Distribution Divergence Measure} \label{subsec:design_d}
Two common families for estimating the divergence between probability distributions are f-divergence (e.g., KL divergence) and integral probability metrics (e.g., Wasserstein distance). In contrast to most work that minimizes the divergence, we need to maximize it, as shown in Eq.~\eqref{eq:max_dom_diff} \&~\eqref{eq:diversity_term}. This strongly suggests to avoid f-divergence because of the near-zero denominators (they tend to generate large but numerically unstable divergence values). Therefore, we choose the second type, specifically the Wasserstein distance, which has been widely used in recent generative modeling methods~\cite{WGAN,SinkhornAutoDiff,CramerDistance,salimans2018improving,SinGAN}.

The Wasserstein distance, also known as optimal transport (OT) distance, is defined as
\begin{equation} \label{eq:wasserstein}
\mathcal{W}_c (\mathbb{P}_a, \mathbb{P}_b) = \inf_{\pi \in \Pi(\mathbb{P}_a, \mathbb{P}_b)} \mathbb{E}_{x_a, x_b \sim \pi} [c(x_a, x_b)],
\end{equation}
where $\Pi(\mathbb{P}_a, \mathbb{P}_b)$ denotes the set of all joint distributions $\pi(x_a, x_b)$ and $c(\cdot, \cdot)$ the transport cost function. Intuitively, the OT metric computes the minimum cost of transporting masses between distributions in order to turn $\mathbb{P}_b$ into $\mathbb{P}_a$.

As the sampling over $\Pi(\mathbb{P}_a, \mathbb{P}_b)$ is intractable, we resort to using the entropy-regularized Sinkhorn distance~\cite{cuturi2013sinkhorn}. Moreover, to obtain unbiased gradient estimators when using mini-batches, we adopt the generalized (squared) energy distance~\cite{salimans2018improving}, leading to
\begin{equation} \label{eq:energy_dist}
d(\mathbb{P}_a, \mathbb{P}_b) = 2\mathbb{E}[\mathcal{W}_c(X_a, X_b)] - \mathbb{E}[\mathcal{W}_c(X_a, X_a^\prime)] - \mathbb{E}[\mathcal{W}_c(X_b, X_b^\prime)],
\end{equation}
where $X_a$ and $X_a^\prime$ are independent mini-batches from distribution $\mathbb{P}_a$; $X_b$ and $X_b^\prime$ are independent mini-batches from distribution $\mathbb{P}_b$; $\mathcal{W}_c$ is the Sinkhorn distance defined as
\begin{equation} \label{eq:sinkhorn}
\mathcal{W}_c(\cdot, \cdot) = \inf_{M \in \mathcal{M}} \sum_{i,j} [M \odot C]_{i,j},
\end{equation}
where the soft-matching matrix $M$ represents the coupling distribution $\pi$ in Eq.~\eqref{eq:wasserstein} and can be efficiently computed using the Sinkhorn algorithm~\cite{SinkhornAutoDiff}; $C$ is the pairwise distance matrix computed over two sets of samples.

Following~\cite{salimans2018improving}, we define the cost function as the cosine distance between instances,
\begin{equation} \label{eq:transport_cost}
c(x_a, x_b) = 1 - \frac{\phi(x_a)^T\phi(x_b)}{||\phi(x_a)||_2 ||\phi(x_b)||_2},
\end{equation}
where $\phi$ is constructed by a CNN (also called critic in~\cite{salimans2018improving}), which maps images into a latent space. In practice, $\phi$ is a fixed CNN that was trained with domain classification loss.

\section{Experiments} \label{sec:experiments}

\subsection{Evaluation on Homogeneous DG}

\begin{figure}[t]
    \centering
    \includegraphics[width=\textwidth]{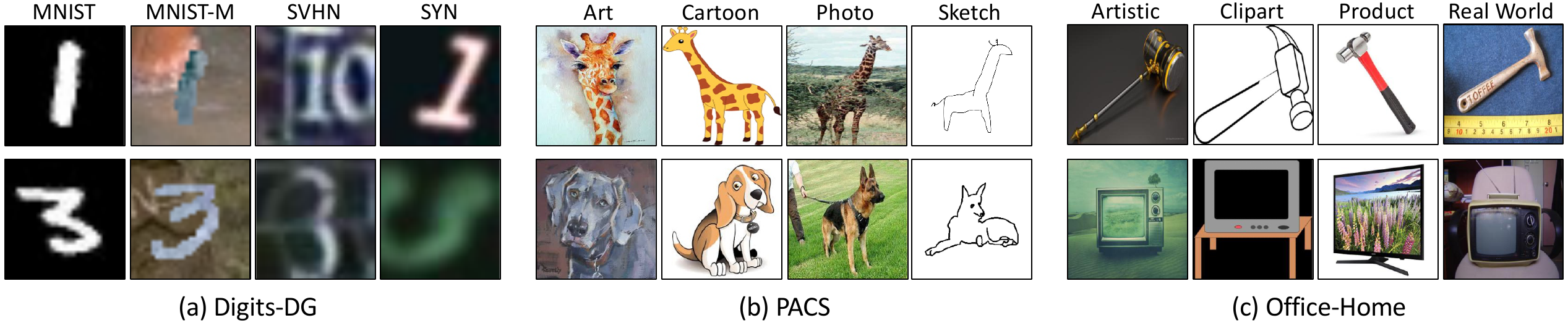}
    \caption{Example images from different DG datasets.}
    \label{fig:example_dataset_images}
\end{figure}

\paragraph{Datasets.}
(1) We use four different digit datasets including MNIST~\cite{lecun1998mnist}, MNIST-M~\cite{ganin2015unsupervised}, SVHN~\cite{netzer2011svhn} and SYN~\cite{ganin2015unsupervised}, which differ drastically in font style, stroke color and background. We call this new dataset \textbf{Digits-DG} hereafter. See Fig.~\ref{fig:example_dataset_images}a for example images.
(2) \textbf{PACS}~\cite{li2017deeper} is composed of four domains, which are Photo, Art Painting, Cartoon and Sketch, with 9,991 images of 7 classes in total. See Fig.~\ref{fig:example_dataset_images}b for example images.
(3) \textbf{Office-Home}~\cite{office_home} contains around 15,500 images of 65 classes for object recognition in office and home environments. It has four domains, which are Artistic, Clipart, Product and Real World. See Fig.~\ref{fig:example_dataset_images}c for example images.

\paragraph{Evaluation protocol.}
For fair comparison with prior work, we follow the leave-one-domain-out protocol in~\cite{li2017deeper,cvpr19JiGen,li2019episodic}. Specifically, one domain is chosen as the test domain while the remaining domains are used as source domains for model training. The top-1 classification accuracy is used as performance measure. All results are averaged over three runs with different random seeds.

\paragraph{Baselines.}
We compare L2A-OT with the recent state-of-the-art DG methods that report results on the same dataset or have code publicly available for reproduction. These include
(1) \textbf{CrossGrad}~\cite{shankar2018generalizing}, the most related work that perturbs input using adversarial gradients from a domain classifier;
(2) \textbf{CCSA}~\cite{motiian2017unified}, which learns a domain-invariant representation using a contrastive semantic alignment loss;
(3) \textbf{MMD-AAE}~\cite{li2018mmdaae}, which imposes a MMD loss on the hidden layers of an autoencoder.
(4) \textbf{JiGen}~\cite{cvpr19JiGen}, which has an auxiliary self-supervision loss to solve the Jigsaw puzzle task~\cite{jigsaw_puzzles};
(5) \textbf{Epi-FCR}~\cite{li2019episodic}, which designs an episodic training strategy;
(6) A \textbf{vanilla} model trained by aggregating all source domains, which serves as a strong baseline.

\begin{table}[t]
\tabstyle{6pt}
\begin{tabular}{l | c c c c | c}
\hline
Method & MNIST & MNIST-M & SVHN & SYN & Avg. \\ \hline
Vanilla & 95.8 & 58.8 & 61.7 & 78.6 & 73.7 \\
CCSA~\cite{motiian2017unified} & 95.2 & 58.2 & 65.5 & 79.1 & 74.5 \\
MMD-AAE~\cite{li2018mmdaae} & 96.5 & 58.4 & 65.0 & 78.4 & 74.6 \\
CrossGrad~\cite{shankar2018generalizing} & \textbf{96.7} & 61.1 & 65.3 & 80.2 & 75.8 \\
JiGen~\cite{cvpr19JiGen} & 96.5 & 61.4 & 63.7 & 74.0 & 73.9 \\
L2A-OT (\emph{ours}) & \textbf{96.7} & \textbf{63.9} & \textbf{68.6} & \textbf{83.2} & \textbf{78.1} \\
\hline
\end{tabular}
\caption{Leave-one-domain-out results on Digits-DG.}
\label{tab:resOnDigitsDG}
\end{table}

\paragraph{Implementation details.}
For Digits-DG, the CNN backbone is constructed with four 64-kernel $3\times3$ convolution layers and a softmax layer. ReLU and $2\times2$ max-pooling are inserted after each convolution layer. $F$ is trained with SGD, initial learning rate of 0.05 and batch size of 126 (42 images per source) for 50 epochs. The learning rate is decayed by 0.1 every 20 epochs. For all experiments, $G$ is trained with Adam~\cite{kingma2014adam} and a constant learning rate of 0.0003. For both PACS and Office-Home, we use ResNet-18~\cite{he2016deep} pretrained on ImageNet~\cite{deng2009imagenet} as the CNN backbone, following~\cite{d2018domain,cvpr19JiGen,li2019episodic}. On PACS, $F$ is trained with SGD, initial learning rate of 0.00065 and batch size of 24 (8 images per source) for 40 epochs. The learning rate is decayed by 0.1 after 30 epochs. On Office-Home, the optimization parameters are similar to those on PACS except that the maximum epoch is 25 and the learning rate decay step is 20. For all datasets, as target data is unavailable during training, the values of hyper-parameters $\lambda_{\mathrm{Domain}}$, $\lambda_{\mathrm{Cycle}}$ and $\lambda_{\mathrm{CE}}$ are set based on the performance on source validation set,\footnote{The searching space is: $\lambda_{\mathrm{Domain}} \in \{0.5, 1, 2\}$, $\lambda_{\mathrm{Cycle}} \in \{10, 20\}$ and $\lambda_{\mathrm{CE}} \in \{1\}$.} which is a strategy commonly adopted in the DG literature~\cite{cvpr19JiGen,li2019episodic}. Our implementation is based on \texttt{Dassl.pytorch}~\cite{zhou2020domain}. 

\paragraph{Results on Digits-DG.}
Table~\ref{tab:resOnDigitsDG} shows that L2A-OT achieves the best performance on all domains and consistently outperforms the vanilla baseline by a large margin. Compared with CrossGrad, L2A-OT performs clearly better on MNIST-M, SVHN and SYN, with clear improvements of 2.8\%, 3.3\% and 3\%, respectively. It is worth noting that these three domains are very challenging with large domain variations compared with their source domains (see Fig.~\ref{fig:example_dataset_images}a). The huge advantage over CrossGrad can be attributed to L2A-OT's unique generation of unseen-domain data using a fully learnable CNN generator, and using optimal transport to explicitly encourage domain divergence. Compared with the domain alignment methods, L2A-OT surpasses MMD-AAE and CCSA by more than 3.5\% on average. The is because L2A-OT enriches the domain diversity of training data, thus reducing overfitting in source domains. L2A-OT clearly beats JiGen because the Jigsaw puzzle transformation does not work well on digit images with sparse pixels~\cite{jigsaw_puzzles}.

\paragraph{Results on PACS.}
The results are shown in Table~\ref{tab:resOnPACS}. Overall, L2A-OT achieves the best performance on all test domains. L2A-OT clearly beats the latest DG methods, JiGen and Epi-FCR. This is because our classifier benefits from the generated unseen-domain data while JiGen and Epi-FCR, like the domain alignment methods, are prone to overfitting to the source domains. L2A-OT beats CrossGrad on all domains, mostly with a large margin. This again justifies our design of learnable CNN generator over adversarial gradient.

\paragraph{Results on Office-Home.}
The results are reported in Table~\ref{tab:resOnOfficeHome}. Again, L2A-OT achieves the best overall performance, and other conclusions drawn previously also hold. Notably, the simple vanilla model obtains strong results on this benchmark, which are even better than most existing DG methods. This is because the dataset is relatively large, and the domain shift is less severe compared with the style changes on PACS and the font variations on Digits-DG.

\begin{table}[t]
\tabstyle{7pt}
\begin{tabular}{l | c c c c | c}
\hline
Method & Art & Cartoon & Photo & Sketch & Avg. \\ \hline
Vanilla & 77.0 & 75.9 & 96.0 & 69.2 & 79.5 \\
CCSA~\cite{motiian2017unified} & 80.5 & 76.9 & 93.6 & 66.8 & 79.4 \\
MMD-AAE~\cite{li2018mmdaae} & 75.2 & 72.7 & 96.0 & 64.2 & 77.0 \\
CrossGrad~\cite{shankar2018generalizing} & 79.8 & 76.8 & 96.0 & 70.2 & 80.7 \\
JiGen~\cite{cvpr19JiGen} & 79.4 & 75.3 & 96.0 & 71.6 & 80.5 \\
Epi-FCR~\cite{li2019episodic} & 82.1 & 77.0 & 93.9 & 73.0 & 81.5 \\
L2A-OT (\emph{ours}) & \textbf{83.3} & \textbf{78.2} & \textbf{96.2} & \textbf{73.6} & \textbf{82.8} \\
\hline
\end{tabular}
\caption{Leave-one-domain-out results on PACS dataset.}
\label{tab:resOnPACS}
\end{table}

\begin{table}[t]
\tabstyle{5pt}
\begin{tabular}{l | c c c c | c}
\hline
Method & Artistic & Clipart & Product & Real World & Avg. \\ \hline
Vanilla & 58.9 & 49.4 & {74.3} & {76.2} & 64.7 \\
CCSA~\cite{motiian2017unified} & {59.9} & {49.9} & 74.1 & 75.7 & {64.9} \\
MMD-AAE~\cite{li2018mmdaae} & 56.5 & 47.3 & 72.1 & 74.8 & 62.7 \\
CrossGrad~\cite{shankar2018generalizing} & 58.4 & 49.4 & 73.9 & 75.8 & 64.4 \\
JiGen~\cite{cvpr19JiGen} & 53.0 & 47.5 & 71.5 & 72.8 & 61.2 \\
L2A-OT  (\emph{ours}) & \textbf{60.6} & \textbf{50.1} & \textbf{74.8} & \textbf{77.0} & \textbf{65.6} \\
\hline
\end{tabular}
\caption{Leave-one-domain-out results on Office-Home.}
\label{tab:resOnOfficeHome}
\end{table}

\begin{table}[t]
\tabstyle{5pt}
\begin{tabular}{l | c c c c | c c c c}
\hline
\multirow{2}{*}{Method} & \multicolumn{4}{c|}{Market1501$\to$Duke} & \multicolumn{4}{c}{Duke$\to$Market1501} \\
 & mAP & R1 & R5 & R10 & mAP & R1 & R5 & R10 \\
\hline
\multicolumn{9}{c}{UDA methods} \\
\hline
ATNet~\cite{liu2019adaptive} & 24.9 & 45.1 & 59.5 & 64.2 & 25.6 & 55.7 & 73.2 & 79.4 \\
CamStyle~\cite{zhong2019camstyle} & 25.1 & \textbf{48.4} & \textbf{62.5} & \textbf{68.9} & 27.4 & 58.8 & 78.2 & \textbf{84.3} \\
HHL~\cite{zhong2018gen} & \textbf{27.2} & 46.9 & 61.0 & 66.7 & \textbf{31.4} & \textbf{62.2} & \textbf{78.8} & 84.0 \\
\hline
\multicolumn{9}{c}{DG methods} \\
\hline
Vanilla & 26.7 & 48.5 & 62.3 & 67.4 & 26.1 & {57.7} & 73.7 & {80.0} \\
CrossGrad~\cite{shankar2018generalizing} & 27.1 & 48.5 & 63.5 & {69.5} & {26.3} & 56.7 & 73.5 & 79.5 \\
L2A-OT (\emph{ours}) & \textbf{29.2} & \textbf{50.1} & \textbf{64.5} & \textbf{70.1} & \textbf{30.2} & \textbf{63.8} & \textbf{80.2} & \textbf{84.6} \\
\hline
\end{tabular}
\caption{Results on cross-domain person re-ID benchmarks.}
\label{tab:resOnXdomReID}
\end{table}

\subsection{Evaluation on Heterogeneous DG}
In this section, we evaluate L2A-OT on a more challenging DG task with disjoint label space between training and test data, namely cross-domain person re-identification (re-ID).

\paragraph{Datasets.}
We use Market1501~\cite{zheng2015scalable} and DukeMTMC-reID (Duke)~\cite{ristani2016perform,zheng2017unlabeled}. Market1501 has 32,668 images of 1,501 identities captured by 6 cameras (domains). Duke has 36,411 images of 1,812 identities captured by 8 cameras.

\paragraph{Evaluation protocol.}
We follow the recent unsupervised domain adaptation (UDA) methods in the person re-ID literature~\cite{zhong2018gen,zhong2019camstyle,liu2019adaptive} and experiment with Market1501$\to$Duke and Duke$\to$Market1501. Different from the UDA setting, we directly test the source-trained model on the target dataset without adaptation. Note that the cross-domain re-ID evaluation involves training a person classifier on source dataset identities. This is then transferred and used to recognize a disjoint set of people in the target domain of unseen camera views via nearest neighbor. Since the label space is disjoint, this is a \emph{heterogeneous} DG problem. For performance measure, we adopt CMC ranks and mAP~\cite{zheng2015scalable}.

\paragraph{Implementation details.}
For the CNN backbone, we employ the state-of-the-art re-ID model, OSNet-IBN~\cite{zhou2019osnet,zhou2019learning}. Following~\cite{zhou2019osnet,zhou2019learning}, OSNet-IBN is trained using the standard classification paradigm, i.e. each identity is considered as a class. Therefore, the entire L2A-OT framework remains unchanged. At test time, feature vectors extracted from OSNet-IBN are used to compute $\ell_2$ distance for image matching. Our implementation is based on \texttt{Torchreid}~\cite{torchreid}.

\paragraph{Results.}
In Table~\ref{tab:resOnXdomReID}, we compare L2A-OT with the vanilla model and CrossGrad, as well as state-of-the-art UDA methods for re-ID. As a result, CrossGrad barely improves the vanilla model while L2A-OT achieves clear improvements on both settings. Notably, L2A-OT is highly competitive with the UDA methods, though the latter make the significantly stronger assumption of having access to the target domain data (thus gaining an unfair advantage). In contrast, L2A-OT generates images of unseen styles (domains) for data augmentation, and such more diverse data leads to learning a better generalizable re-ID model.

\begin{figure}[t]
    \centering
    \includegraphics[width=\textwidth]{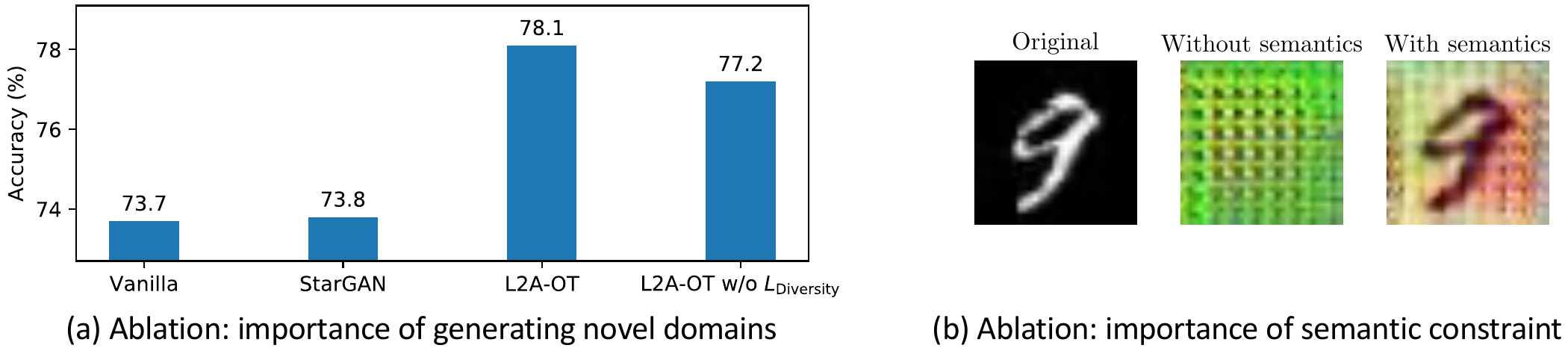}
    \caption{Ablation study.}
    \label{fig:ablation}
\end{figure}

\subsection{Ablation Study}

\paragraph{Importance of generating novel domains.}
To verify that our improvement is brought by the increase in training data distributions by the generated novel domains (i.e. Eq.~\eqref{eq:max_dom_diff} \&~\eqref{eq:diversity_term}), we compare L2A-OT with StarGAN~\cite{StarGAN}, which generates data from the existing source domains by performing source-to-source mapping. The experiment is conducted on Digits-DG and the average performance over all test domains is used for comparison. Fig.~\ref{fig:ablation}a shows that StarGAN performs only similarly to the vanilla model (StarGAN's 73.8\% vs. vanilla's 73.7\%) while L2A-OT obtains a clear improvement of 4.3\% over StarGAN. This confirms that increasing domains is far more important than increasing data (of seen domains) for DG. Note that this 4.3\% gap is attributed  to the combination of the OT-based domain novelty loss (Eq.~\eqref{eq:max_dom_diff}) and the diversity loss (Eq.~\eqref{eq:diversity_term}). 
Fig.~\ref{fig:ablation}a shows that the diversity loss  contributes around 1\% to the performance, and the rest improvement comes from the diversity loss.

\paragraph{Importance of semantic constraint.}
The cycle-consistency and cross-entropy losses (Eq.~\eqref{eq:cycle} \&~\eqref{eq:cross_entropy}) are essential in the L2A-OT framework for maintaining the semantic content when performing domain translation. Fig.~\ref{fig:ablation}b shows that without the semantic constraint, the content is completely missing (we found that using these images reduced the result from 78.1\% to 73.9\%).

\subsection{Further Analysis}

\begin{figure}[t]
    \centering
    \includegraphics[width=.65\textwidth]{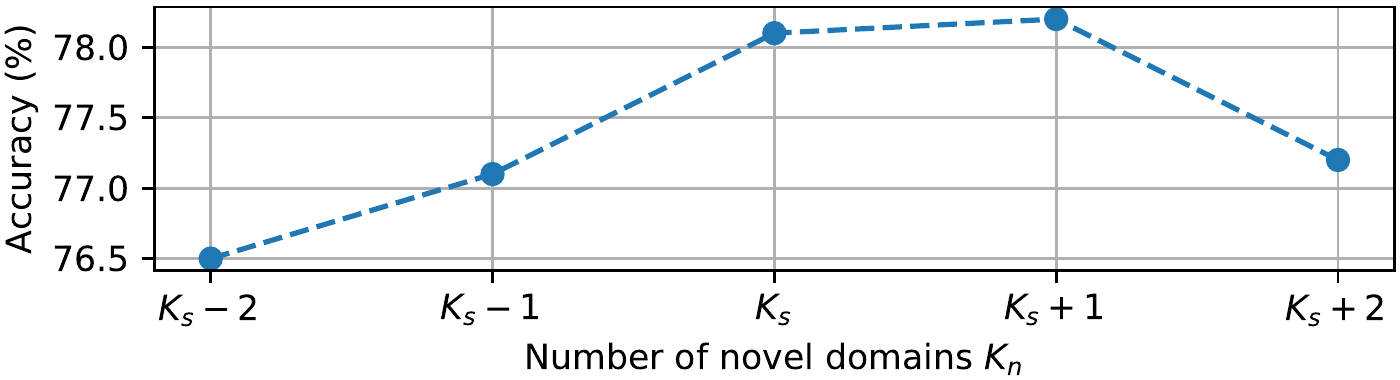}
    \caption{Results of varying $K_n$. Here $K_s=3$.}
    \label{fig:num_novel_domains}
\end{figure}

\paragraph{How many novel domains to generate?}
Our approach can generate an arbitrary number of novel domains, although we have always doubled the number of domains (set $K_s=K_n$) so far. Fig.~\ref{fig:num_novel_domains} investigates the significance on the choice of number of novel domains. In principle, synthesizing more domains provides opportunity for more diverse data, but also increases optimization difficulty and is dependent on the source domains. The result shows that the performance is not very sensitive to the choice of novel domain number, with $K_n=K_s$ being a good rule of thumb.

\begin{table}[t]
\tabstyle{8pt}
\begin{tabular}{c c c | c | c c}
\hline
\multicolumn{3}{c|}{Source} & \multirow{2}{*}{Target} & \multirow{2}{*}{L2A-OT} & \multirow{2}{*}{Vanilla} \\
MNIST & SVHN & SYN & & & \\
\hline
$\checkmark$ & $\checkmark$ & & MNIST-M & 60.9 & 54.6 \\
$\checkmark$ & & $\checkmark$ & MNIST-M & 62.1 & \textbf{59.1} \\
 & $\checkmark$ & $\checkmark$ & MNIST-M & 49.7 & 45.2 \\
\hline
$\checkmark$ & $\checkmark$ & $\checkmark$ & MNIST-M & \textbf{62.5} & 57.1 \\
\hline
\end{tabular}
\caption{Using two vs. three source domains on Digits-DG where the size of training data is kept identical for all settings for fair comparison.}
\label{tab:two_vs_three_src_dom}
\end{table}

\begin{figure}[t]
    \centering
    \includegraphics[width=\textwidth]{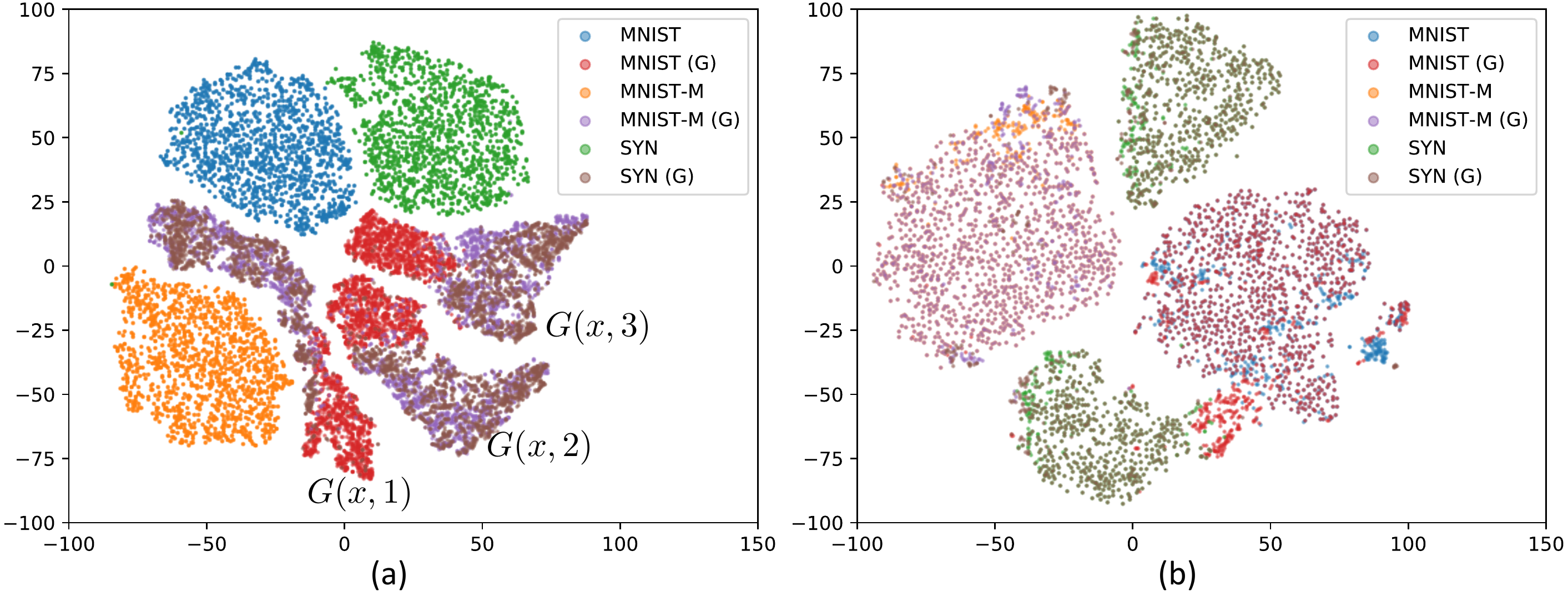}
    \caption{T-SNE visualization of domain embeddings of (a) L2A-OT and (b) CrossGrad~\cite{shankar2018generalizing}. X (G) indicates novel data when using the domain X as a source.\cut{In (a), $G(x, i)$ means the $i$-th novel domain ($x$ comes from the three sources).}}
    \label{fig:domain_distribution_vs_crossgrad}
\end{figure}

\begin{figure}[t]
    \centering
    \includegraphics[width=\textwidth]{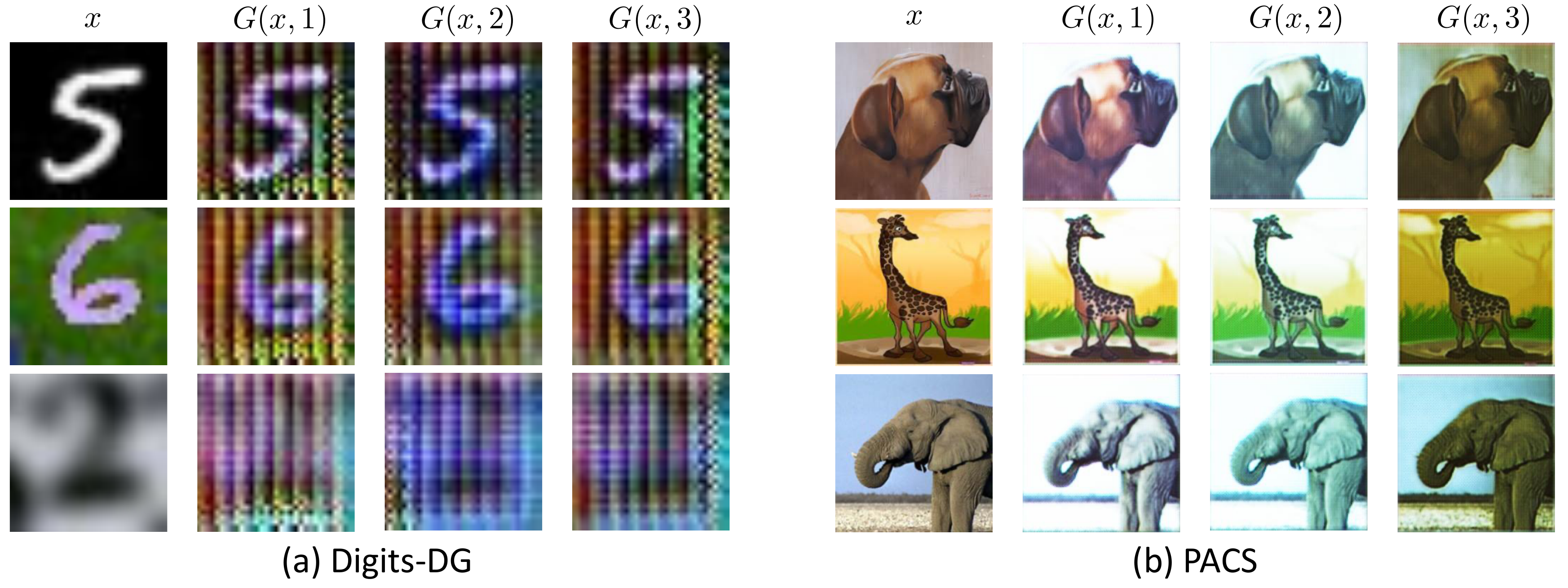}
    \caption{Visualization of generated images. $x$: source image. $G(x, i)$: generated image of the $i$-th novel domain.}
    \label{fig:vis_g_output}
\end{figure}

\begin{figure}[t]
    \centering
    \includegraphics[width=.9\textwidth]{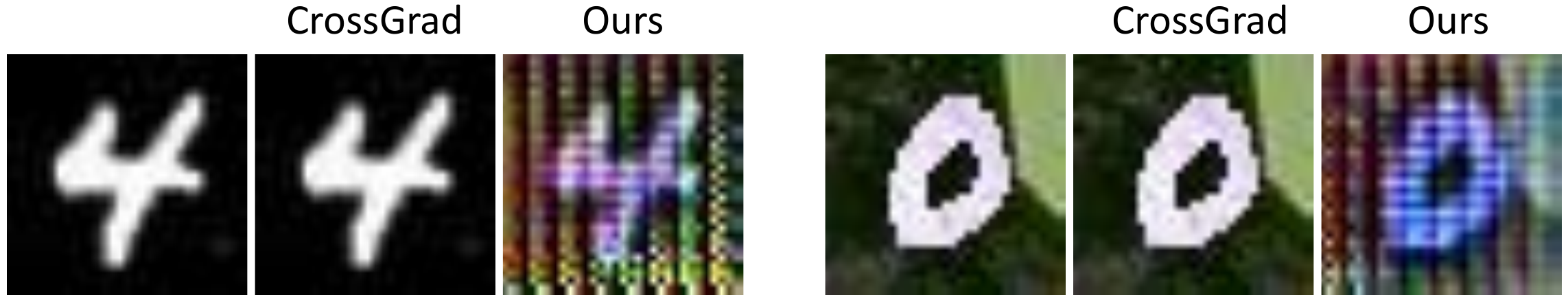}
    \caption{Comparison between L2A-OT and CrossGrad~\cite{shankar2018generalizing} on image generation.}
    \label{fig:vis_g_vs_crossgrad}
\end{figure}

\paragraph{Do more source domains lead to a better result?}
In general, yes. The evidence is shown in Table~\ref{tab:two_vs_three_src_dom} where the result of using three sources is generally better than using two as we might expect due to the additional diversity. The detailed results show that when using two sources, performance is sensitive to the choice of sources among the available three. This is expected since different sources will vary in transferrability to a given target. However, for both vanilla and L2A-OT the performance of using three sources is better than the performance of using two averaged across the 2-source choices.

\paragraph{Visualizing domain distributions.}
We employ t-SNE~\cite{tsne} to visualize the domain feature embeddings using the validation set of Digits-DG (see Fig.~\ref{fig:domain_distribution_vs_crossgrad}a).
We have the following observations.
(1) The generated distributions are clearly separated from the source domains and evenly fill the unseen domain space.
(2) The generated distributions form independent clusters (due to our diversity term in Eq.~\eqref{eq:diversity_term}).
(3) $G$ has successfully learned to flexibly transform one source domain to any of the discovered novel domains.

\paragraph{Visualizing novel-domain images.}
Fig.~\ref{fig:vis_g_output} visualizes the output of $G$. In general, we observe that the generated images from different novel domains manifest different properties and more importantly, are clearly different from the source images. For example, in Digits-DG (Fig.~\ref{fig:vis_g_output}a), $G$ tends to generate images with different background patterns/textures and font colors. In PACS (Fig.~\ref{fig:vis_g_output}b), $G$ focuses on contrast and color. Fig.~\ref{fig:vis_g_output} seems to suggest that the synthesized domains are not drastically different from each other. However,  a seemingly limited
diversity in the image space to human eyes can be significant
to a CNN classifier: both Fig.~\ref{fig:main_idea} and Fig.~\ref{fig:domain_distribution_vs_crossgrad}a show clearly that the synthesized
data points have very different distributions from both
the original ones and each other in a feature  embedding space, making them useful for
learning a domain-generalizable classifier.

\paragraph{L2A-OT vs.~CrossGrad.}
It is clear from Fig.~\ref{fig:domain_distribution_vs_crossgrad}b that the new domains generated by CrossGrad largely overlap with the original domains. This is because CrossGrad is based on adversarial attack methods~\cite{goodfellow2015explaining}, which are designed to make imperceptible changes. This is further verified in Fig.~\ref{fig:vis_g_vs_crossgrad} where the images generated by CrossGrad have only subtle differences in contrast to the original images. On the contrary, L2A-OT can model much more complex domain variations that can materially benefit the classifier, thanks to the full CNN image generator and OT-based domain divergence losses.

\section{Conclusion} \label{sec:conclusion}
We presented L2A-OT, a novel data augmentation-based DG method that boosts classifier's robustness to domain shift by learning to synthesize images from diverse unseen domains through a conditional generator network. The generator is trained by maximizing the OT distance between source domains and pseudo-novel domains. Cycle-consistency and classification losses are imposed on the generator to further maintain the structural and semantic consistency during domain translation. Extensive experiments on four DG benchmark datasets covering a wide range of visual recognition tasks demonstrate the effectiveness and versatility of L2A-OT.

%
%
\bibliographystyle{splncs04}
\bibliography{egbib}
\end{document}